%% file: arxivV2.tex
\documentclass[times, twoside, watermark]{zreview-styletex}
\usepackage{blindtext}
\usepackage{cuted}
\usepackage[hang,flushmargin]{footmisc} 
\pgfplotsset{compat=1.16}

\newcounter{texexp}
\tcbset{
texexp/.style={colframe=myblue!80!black, colback=myblue!5!white,
fonttitle=\small\sffamily\bfseries},
example/.code 2 args={\refstepcounter{texexp}\label{#2}%
\pgfkeysalso{texexp,title={Box \thetexexp: #1}}},
}

\usepackage[para]{threeparttable}

\leadauthor{Stoffl et al.} 
\usepackage{graphicx}

\usepackage{tikz}
\usepackage{pgfplots}
\pgfplotsset{compat=1.14}

\definecolor{curveblue}{rgb}{0.1725490196,0.74901960784,0.96862745098}
\definecolor{curvered}{rgb}{0.98823529411, 0.34117647058, 0.05882352941}
\definecolor{curvegreen}{rgb}{0.18823529411, 0.974117647058, 0.2282352941}
\definecolor{curvemap}{rgb}{0.996078, 0.3725490196, 0.333}

\usepackage{epstopdf}

\usepackage{dsfont}
\DeclareMathOperator*{\argmin}{arg\,min}

\usepackage{booktabs}
\usepackage{enumitem}
\usepackage{verbatim}
\usepackage{dblfloatfix}

\usepackage{lipsum,lmodern}
\usepackage[most]{tcolorbox}
\usepackage{array}

\begin{document}

\title{End-to-End Trainable Multi-Instance Pose Estimation with Transformers}
\shorttitle{POET}
\author{Lucas Stoffl ~~~~~~~~~~~~~~ Maxime Vidal ~~~~~~~~ Alexander Mathis \\
\vspace{0.2em}
{\normalsize Ecole Polytechnique Fédérale de Lausanne (EPFL);
CH-1015 Lausanne, Switzerland} \\
\vspace{0.4em}
{\tt\small alexander.mathis@epfl.ch}
\vspace{2em}
}

\maketitle

\begin{abstract}
We propose an end-to-end trainable approach for multi-instance pose estimation, called POET ({\bf PO}se {\bf E}stimation {\bf T}ransformer). Combining a convolutional neural network with a transformer encoder-decoder architecture, we formulate multi-instance pose estimation from images as a direct set prediction problem. Our model is able to directly regress the pose of all individuals, utilizing a bipartite matching scheme. POET is trained using a novel set-based global loss that consists of a keypoint loss, a visibility loss and a class loss. POET reasons about the relations between multiple detected individuals and the full image context to directly predict their poses in parallel. We show that POET achieves high accuracy on the challenging COCO keypoint detection task while having less parameters and higher inference speed than other bottom-up and top-down approaches. Moreover, we show successful transfer learning when applying POET to animal pose estimation. To the best of our knowledge, this model is the first end-to-end trainable multi-instance pose estimation method and we hope it will serve as a simple and promising alternative.
\end{abstract}

\section*{Introduction}

\indent Multi-instance pose estimation from a single image, the task of predicting the body part locations for each individual, is an important computer vision problem. It has wide ranging applications from measuring behavior in health care and biology to virtual reality and human-computer interactions~\cite{Poppe2007surveypose, mathis2018deeplabcut, chen2020monocular, mathis2020primer,uchida2021biomechanics}.

Multi-human pose estimation can be thought of as a hierarchical set prediction task. An algorithm needs to predict the bodyparts of all individuals and group them correctly into humans, i.e., it needs to predict a set of (bodypart) sets (per individual). Due to the complexity of this process, current methods {\it consist of multiple steps and are not end-to-end trainable}. Fundamentally, top-down and bottom-up methods are the major approaches. Top-down methods first predict the location (bounding boxes) of all individuals based on an object detection algorithm and then predict the location of all the bodyparts per cropped individual with a separate network~\cite{xiao2018simple, sun2019deep,zhang2020distribution,yang2020transpose} -- thus one needs $n+1$ network evaluations per image with $n$ individual (detections). 
Bottom-up methods first predict all the bodyparts, and then group them into individuals with various techniques~\cite{cao2017realtime,insafutdinov2017arttrack, papandreou2017towards, newell2017associative, kreiss2019pifpaf, nie2019single, cheng2020higherhrnet, kreiss2021openpifpaf}. Hence, both approaches require either two different networks or complex post-processing. This motivates the search for end-to-end solutions.

\input{fig_arxiv2/inference_time_fps_woDef}

\begin{figure*}
  \centering
  \includegraphics[width=.9\linewidth]{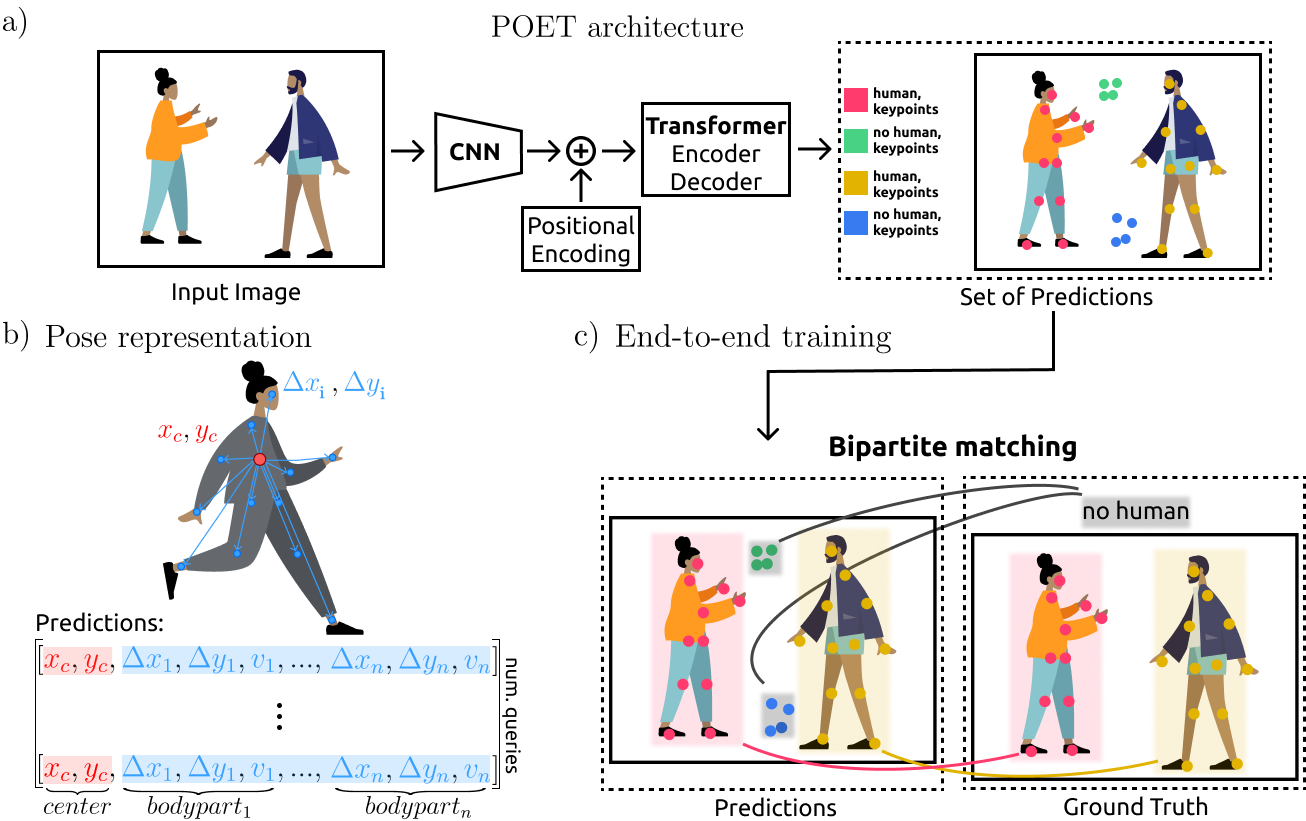}  \caption{Overview of our model. {\bf a)} POET combines a CNN backbone and a transformer to directly predict the pose of multiple humans. {\bf b)} Each pose is represented as a vector comprising the center $(x_c, y_c)$, the relative offset $(\Delta x_i, \Delta y_i)$ of each bodypart $i$ and its visibility $v_i$. {\bf c)} POET is trained end-to-end by bipartite matching of the closest predictions to the ground truth pose, and then backpropagating the loss.}
  \label{fig:idea}
\end{figure*}

Inspired by DETR~\cite{carion2020end}, a recent transformer-based architecture for object detection, we propose a novel end-to-end trainable method for multi-instance pose estimation. \textbf{PO}se \textbf{E}stimation \textbf{T}ransformer (POET) is the first model that is trained end-to-end for multi-instance pose estimation, without the need for any post-processing or two networks as typically employed in top-down approaches. POET is trained with a novel, yet simple loss function, which allows bipartite matching between predicted and ground-truth human poses. Our approach achieves strong results on the COCO keypoint challenge~\cite{lin2014microsoft}, especially for large humans, and performs better than a baseline model even with higher spatial resolution.
Analyses of the transformer attentions and of the decoder depth give insight into how POET tackles the problem of multi-instance pose estimation, while extensive loss ablations show the importance of computing an adaptable center that is successfully learned by POET.
Moreover, we show that POET also achieves strong performance on a much smaller dataset for animal pose estimation (MacaquePose~\cite{labuguen2020macaquepose}) for which the same COCO-defined keypoints have been annotated. We show that the transformer weights can be transferred from COCO to MacaquePose, yielding even better performance.
At last, POET shows excellent inference speed, being faster than various light-weight bottom-up and top-down methods (Figure~\ref{fig:speed_fps_wo}).

\section*{Related Work}

\subsection*{Transformers in Vision and Beyond:}

Transformers were first introduced for machine translation~\cite{vaswani2017attention}, and have vastly improved the performance of deep learning models on language tasks~\cite{vaswani2017attention,lan2019albert,brown2020language}. Their architecture inherently allows modeling and discovering long-range interactions in data and their use has recently been extended to speech recognition~\cite{dong2018speech}, automated theorem proving~\cite{polu2020generative}, and many other tasks~\cite{khan2021transformers}. In computer vision, transformers have been used with great effect either in combination or as an alternative to convolutional neural networks (CNNs)~\cite{khan2021transformers}. Notably, Visual Transformer (ViT)~\cite{dosovitskiy2020image} demonstrated state-of-the-art performance on image recognition tasks with pure transformer models. In other visual tasks, such as text-to-image, excellent results have been shown, e.g., by DALL-E~\cite{ramesh2021zero}.

Recently Carion et al.~\cite{carion2020end} developed a new end-to-end paradigm for visual object detection with transformers, a task which previously required either two-stage approaches or post-processing. This approach, DETR, formulated object detection as a set prediction problem combined with a bipartite matching loss. DETR is an elegant solution, however, the model requires long training times and shows comparatively low performance on small objects~\cite{carion2020end}. These problems were mitigated through further works; Deformable DETR~\cite{zhu2020deformable} presents a multi-scale deformable attention module which only attends to a set number of points within the feature map, for different scales, and in this way reducing the training time and improving small object detection performance. Sun et al. removed the transformer decoder and fed the features coming out of the CNN backbone to a Feature Pyramid Network~\cite{sun2020rethinking}.

Importantly, end-to-end approaches were successfully applied in many complex prediction tasks such as speech recognition or machine translation~\cite{lan2019albert,lan2019albert,brown2020language}, but are still {\it lacking in multi-instance pose estimation}.

\subsection*{Pose Estimation} is a classical computer vision problem with wide ranging applications~\cite{Poppe2007surveypose, chen2020monocular,zheng2020deep, mathis2020primer}. Pose estimation methods are evaluated on several benchmarks for 2D multi-human pose estimation, incl. COCO~\cite{lin2014microsoft,andriluka20142d,li2019crowdpose,chen2020monocular}

\input{fig_arxiv2/losses}

\begin{figure*}
  \centering
  \includegraphics[width=\linewidth]{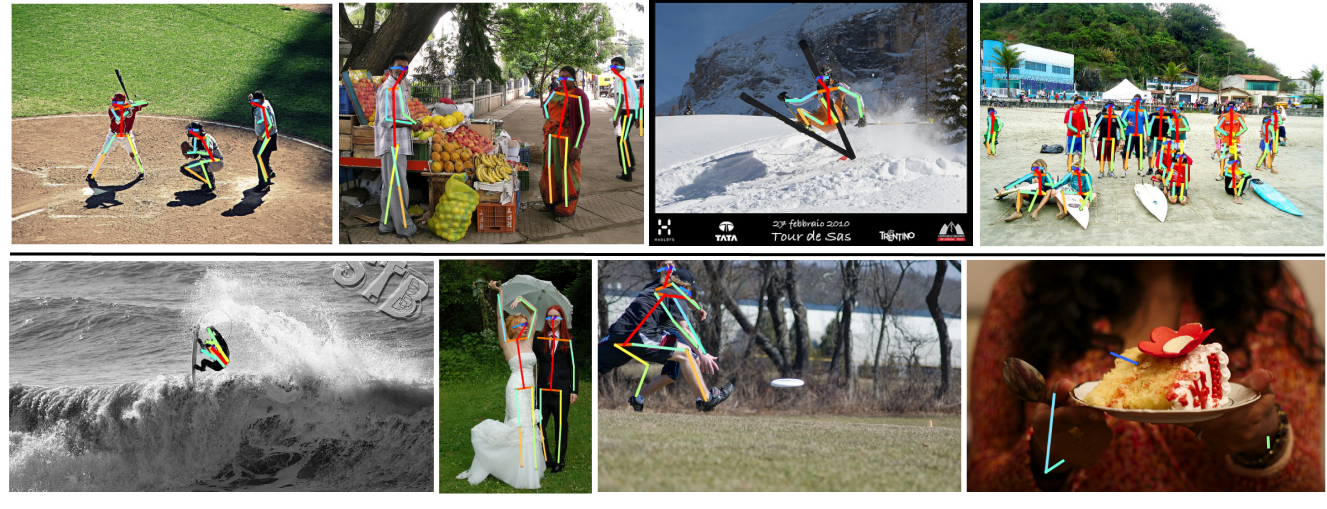}
  \caption{Example predictions on COCO-eval with POET-R50 model listed in Table~\ref{table_results_sota}. Top row: examples with good performance. Bottom row: examples with errors.}
    \label{fig:example_imgs}
\end{figure*}

\begin{table*}[ht]
\centering
\vspace{15pt}
\caption{Comparison to state-of-the-art models on COCO test-dev. Note that most models use an overall stride of 4 (or smaller), when extracting features, while POET uses stride 32 and therefore much smaller feature maps which harms performance on AP$_{M}$. However, with regard to AP$_{L}$ it can compete with state-of-the-art models.}
\label{table_results_sota} \footnotesize
\begin{tabular}{l|c|c|ccccc|ccccc}
\toprule

Method & Stride & \#Params &  AP & AP$_{50}$ & AP$_{75}$ & AP$_{M}$ & AP$_{L}$ & AR & AR$_{50}$ & AR$_{75}$ & AR$_{M}$ & AR$_{L}$ \\
\toprule
\multicolumn{2}{l}{Top-Down}\\
\midrule
Mask-RCNN~\cite{he2017mask} & 4 & 44.4M & 63.1 & 87.3 & 68.7 & 57.8 & 71.4 & - & - & - & - & - \\
CPN~\cite{chen2018cascaded} & - & - & 72.1 & 91.4 & 80.0 & 68.7 & 77.2 & 78.5 & 95.1 & 85.3 & 74.2 & 84.3 \\
HRNet-W48~\cite{sun2019deep, wang2020deep} & 4 & 63.6M & {\bf 75.5} & 92.5 & 83.3 & 71.9 & 81.5 & 80.5 & - & - & - & - \\
 
TansPose-H-A6~\cite{yang2020transpose} & 4 & 17.5M &  75.0 & 92.2 & 82.3 & 71.3 & 81.1 & - & - & - & - & - \\
\midrule
\multicolumn{2}{l}{Bottom-Up}\\
\midrule
OpenPose~\cite{cao2017realtime} & 4 & - & 61.8 & 84.9 & 67.5 & 57.1 & 68.2 & - & - & - & - & - \\
Hourglass~\cite{newell2017associative} & 4 & 277.8M & 56.6 & 81.8 & 61.8 & 49.8 & 67.0 & - & - & - & - & - \\
PersonLab~\cite{papandreou2018personlab} & 8 & 68.7M & 66.5 & 88.0 & 72.6 & 62.4 & 72.3 & 71.0 & 90.3 & 76.6 & 66.1 & 77.7 \\
AE + R50~\cite{mmpose2020aeresnet} & 4 & 31.9M & 46.6 & 74.2 & 47.9 & 44.6 & 49.3 & 55.2 & 79.7 & 57.5 & 48.1 & 65.1 \\
PifPaf~\cite{kreiss2019pifpaf} & 4 & - & 66.7 & - & - & 62.4 & 72.9 & - & - & - & - & - \\
HigherHRNet~\cite{cheng2020higherhrnet} & 2 & 63.8M & 68.4 & 88.2 & 75.1 & 64.4 & 74.2 & - & - & - & - & - \\
\textbf{POET-R50 (Ours)} & 32 & 41.3M & 53.6 & 82.2 & 57.6 & 42.5 & 68.1 & 61.4 & 87.8 & 65.5 & 50.8 & 75.7 \\
\bottomrule
\end{tabular}
\end{table*}

Multi-instance pose estimation methods can be classified as top-down and bottom-up~\cite{chen2020monocular,zheng2020deep, mathis2020primer}. Top-down methods predict the location of each bodypart of each {\it individual}, based on bounding boxes localizing individuals with a separate network~\cite{newell2016stacked,iqbal2016multi, xiao2018simple,chen2018cascaded, sun2019deep,zhang2020distribution,yang2020transpose}. Bottom-up methods first predict all the bodyparts, and then group them into individuals by using part affinity fields~\cite{cao2017realtime}, pairwise predictions~\cite{insafutdinov2016deepercut,insafutdinov2017arttrack, papandreou2017towards, Dang2019,nie2019single},  composite fields~\cite{kreiss2021openpifpaf, kreiss2019pifpaf}, or associative embeddings~\cite{newell2017associative, cheng2020higherhrnet}. Both top-down and bottom-up approaches, require either post-processing steps (for assembly) or two different neural networks (for localization and then pose estimation). Thus, these multi-stage approaches are prone to errors and typically show poor inference speed.

Most recent (state-of-the-art) approaches are fully convolutional and predict keypoint heatmaps (e.g.,~\cite{ sun2019deep,zhang2020distribution,newell2017associative,cao2017realtime, cheng2020higherhrnet}). Recently, Yang et al. proposed TransPose, a top-down method, which predicts heatmaps as well, but by using attention after a CNN encoder~\cite{yang2020transpose}. Mao et al. applies a top-down transformer architecture with the difference that it is directly regresses the poses (for single persons)~\cite{mao2021tfpose}. A regression-based method is also proposed by the authors of~\cite{li2021pose} building a two-stage network consisting of a person-detection transformer (pre-trained DETR~\cite{carion2020end}) and a keypoint-detection transformer. Transformers were also used for 3D pose and mesh reconstruction on (single) humans, which achieved state-of-the-art on Human3.6M~\cite{zheng20213d, lin2020METRO}.
Extending this work, we build on DETR~\cite{carion2020end}, to propose an end-to-end trainable pose estimation method for multiple instances that directly outputs poses as vectors (without relying on any heatmaps). To cast pose estimation as a hierarchical set prediction problem, we adapt the pose representations of CenterNet~\cite{Dang2019} and Single-Stage Multi-Person Pose Machines~\cite{nie2019single}.

\section*{The POET Model}

Our work is closely related to DETR~\cite{carion2020end} and fundamentally extends this object detection framework to multi-instance pose estimation by introducing a novel loss function and a pose representation that, combined, can tackle the difficulties faced in multi-instance pose estimation, such as non-visible bodyparts. POse Estimation Transformer (POET) consists of two major ingredients: (1) a {\it transformer-based architecture} that predicts a set of human poses in parallel (Figure~\ref{fig:idea}a) and (2) a {\it set prediction loss} that is a linear combination of simple sub-losses for classes, keypoint coordinates and visibilities. To cast multi-instance pose estimation as a set prediction problem, we represent the pose of each individual as the center (of mass) together with the relative offsets per bodypart. Each bodypart can be occluded or visible. POET is trained to directly output a vector comprising the center, relative bodyparts as well as (binary) bodypart visibility indicators (Figure~\ref{fig:idea}b). The POET architecture contains three main elements: a CNN backbone that extracts features of the input images, an encoder-decoder transformer and a pose prediction head that outputs the set of estimated poses.

\subsubsection*{CNN backbone}

The convolutional backbone is given a batch of images, $I \in \mathbb{R}^{B \times 3 \times H\times W}$, as input with batch size $B$, 3 color channels and image dimensions $(H, W)$. Through several computing and downsampling steps, the CNN generates lower-resolution feature maps, $F \in \mathbb{R}^{B \times C \times H/S \times W/S}$ with stride $S$. Specifically, we choose different ResNets~\cite{he2016deep} with various strides $S$, as detailed in the experiments section.

\subsubsection*{Encoder-Decoder transformer}

The encoder-decoder transformer model follows the standard design~\cite{vaswani2017attention,carion2020end}. Both encoder and decoder consist of 6 layers with 8 attention heads each. The encoder takes the output features of the CNN backbone, reduces their channel dimensions by a $1 \times 1$ convolution and collapses them along the spatial dimension into one dimension as the multi-head mechanism expects sequential input. It contains a \textit{fixed} positional encoding to the encoder input, as the transformer architecture is (otherwise) permutation-invariant and would disregard the spatial image structure. In contrast, the input embeddings for the decoder are \textit{learned} positional encodings,  which we refer to as {\it object queries}. They are added to the encoder output to form the decoder input. The decoder transforms the queries into output embeddings, which are then taken by the {\it pose prediction head} and independently decoded into the final set of poses and class labels. Thereby, every query can search for one instance and predicts its pose and class. With the aid of the self-attention in both the encoder and the decoder, the network is able to globally reason about all objects together using pairwise relations between them, and at the same time using the whole image as context information.

\subsubsection*{Pose prediction head}

The $N$ query embeddings at the end of the transformer decoder are independently processed by a feedforward network. This pose estimation head consists of (1) a 3-layer perceptron with ReLU activation, which transforms the embeddings into normalized (with respect to the image size) center coordinates, the displacements to all bodyparts relative to the center and the visibility scores for every body part in a single vector (Figure~\ref{fig:idea}b); and (2) a linear projection layer predicting the class label using a softmax function.

\subsection*{Training Loss:}

In order to predict all poses in parallel, the network is trained with the loss after finding an optimal matching between predictions and ground-truth and summing over the individuals. Therefore, our loss has to score predictions accordingly, with respect to the class, the keypoint coordinates and their visibilities.

From the absolute joint coordinates $A_i=[(x_1, y_1), (x_2, y_2), \dots , (x_n, y_n)]$ for every instance {\it $i$} in the ground truth we compute the center as the center of mass of all visible keypoints. The ground truth vector for individual $i$ is then $[x_c, y_c, \Delta x_1, \Delta y_1, v_1, \Delta x_2, \Delta y_2, v_2 \dots , \Delta x_n, \Delta y_n, v_n]$, for center $(x_c,y_c)$, relative offset $(\Delta x_i, \Delta y_i)$ of each bodypart $i$ and its visibility $v_i$. In order to make the loss functions more legible, we split this vector into $y_i=(c_i,C_i, Z_i, V_i)$, which consists of the target class label (individuals/non-object) $c_i$, the center $C_i= (x_c,y_c)$, the relative pose: $Z_i=[\Delta x_1, \Delta y_1, \Delta x_2, \Delta y_2, \dots , \Delta x_n, \Delta y_n]$ (relative joint displacements from the center $C_i$) and a binary visibility vector $V_i=[v_1, v_1, v_2, v_2, \dots , v_n, v_n]$ encoding for every joint in the image, whether it is visible or not.

The prediction of the network for instance $i$ is then defined as $\widehat{y}_i=(\widehat{p}(c_i), \widehat{C}_i, \widehat{Z}_i, \widehat{V}_i)$, where $\widehat{p}(c_i)$ is the predicted probability for class $c_i$, $\widehat{C}_i$ the predicted center, $\widehat{Z}_i$ the predicted pose, and $\widehat{V}_i$ the predicted visibility. Note that the network does not predict the visibility for the center. By adding the predicted offsets to the predicted centers we obtain the predicted, absolute keypoint positions: $\widehat{A}_i=[(\widehat{x_c}+\widehat{\Delta x_1}, \widehat{y_c}+\widehat{\Delta y_1}), (\widehat{x_c}+\widehat{\Delta x_2}, \widehat{y_c}+\widehat{\Delta y_2)}, \dots , (\widehat{x_c}+\widehat{\Delta x_n}, \widehat{y_c}+\widehat{\Delta y_n})]$.

In the following, we denote by $y$ the ground truth set of poses, and $\widehat{y} = \{\widehat{y}_i\}^N_{i=1}$ the set of $N$ predictions. Here, $y$ is the set of individuals in the image padded with non-objects. We define our pair-wise matching cost between ground truth $y_i$ and a prediction with index $\sigma(i)$ as:

\begin{align} \label{eq:match}
&\mathcal{L}_{\mathrm{match}}(y_i, \widehat{y}_{\sigma(i)})) = -\mathds{1}_{\{c_i \neq \varnothing\}} \widehat{p}_{\sigma(i)}(c_i) \nonumber \\
&+ \mathds{1}_{\{c_i \neq \varnothing\}} \mathcal{L}_{\mathrm{pose}}(C_i, Z_i, V_i, \widehat{C}_{\sigma}(i), \widehat{Z}_{\sigma}(i), \widehat{V}_{\sigma}(i))
\end{align}
Here, $\mathcal{L}_{\mathrm{pose}}$ is the pose-specific cost that we will define below and involves costs for the centers, the bodyparts and their visibilities.

The optimal assignment is then found as a bipartite matching with the lowest matching cost based on the Hungarian algorithm~\cite{kuhn1955hungarian, carion2020end}. This assignment is the following permutation of $N$ elements $\sigma \in {\mathfrak{G}_N}$ for symmetric group ${\mathfrak{G}_N}$~\cite{van2003algebra}: 
\begin{equation} \label{eq:bipartite}
\widehat{\sigma} = \argmin_{\sigma \in \mathfrak{G}_N} \sum_{i}^{N} \mathcal{L}_{\mathrm{match}}(y_i, \widehat{y}_{\sigma(i)})
 \end{equation}

Once the optimal matching is obtained, we can compute the \textit{Hungarian loss} for all matched pairs. Like the matching cost, it contains a loss part scoring the poses, which is a linear combination of a $L_1$ loss to compute the differences between absolute keypoint coordinates, a $L_2$ loss for the visibilities and two regularization terms, namely a $L_2$ loss for the center coordinates and a $L_1$ loss for the keypoint offsets. Thus we have coefficients $\lambda_{abs}$, $\lambda_{vis}$, $\lambda_{ctr}$ and $\lambda_{deltas}$:
\begin{align} \label{eq:pose}
\mathcal{L}_{\mathrm{pose}}&(C_i, Z_i, V_i, \widehat{C}_{\sigma}(i), \widehat{Z}_{\sigma}(i), \widehat{V}_{\sigma}(i)) = \nonumber \\
& \lambda_{abs}\|V_i \circ A_i - V_i \circ \widehat{A}_{\sigma(i)}\|_1 \nonumber \\
&+ \lambda_{vis}\|V_i-\widehat{V}_{\sigma(i)}\|_2^2 \nonumber \\
&+ \lambda_{ctr}\|(x_c, y_c)_i-(\widehat{x_c},\widehat{y_c})_{\sigma(i)}\|_2^2 \nonumber \\
&+ \lambda_{deltas}\|V_i \circ Z_i - V_i \circ \widehat{Z}_{\sigma(i)}\|_1
\end{align}
Thereby, $\circ$ denotes point-wise multiplication. These four losses are normalized by the number of ground truth individuals per batch.

The final loss, the \textit{Hungarian loss}, then is a linear combination of a negative log-likelihood for class prediction and the keypoint-specific loss defined above, for all pairs from the optimal assignment $\widehat{\sigma}$:
\begin{align} \label{eq:hungarian}
&\mathcal{L}_{\mathrm{Hungarian}}(y,\widehat{y}) = \sum_{i=1}^{N} \Big[-\log\widehat{p}_{\widehat{\sigma}(i)}(c_i) \\
&+ \mathds{1}_{\{c_i \neq \varnothing\}} \mathcal{L}_{\mathrm{pose}}(C_i, Z_i, V_i, \widehat{C}_{\widehat{\sigma}}(i), \widehat{Z}_{\widehat{\sigma}}(i), \widehat{V}_{\widehat{\sigma}}(i))\Big] \nonumber
\end{align}

\section*{Experiments}

We evaluated POET on the COCO keypoint estimation challenge~\cite{lin2014microsoft} and MacaquePose~\cite{labuguen2020macaquepose}. We illustrate qualitative results and show that it reaches good performance (especially for large individuals). Then we show that it outperforms baseline methods that we trained based on an established bottom-up method using associative embedding with the same backbone~\cite{newell2017associative, mmpose2020}. Then, we analyze different aspects of the architecture and the loss. Finally, we discuss challenges and future work.

\subsection*{COCO Keypoint Detection Challenge:}

The COCO dataset~\cite{lin2014microsoft} comprises more than $200,000$ images with more than $150,000$ people for which up to $17$ keypoints are annotated. The dataset is split into train/val/test-dev sets with $57k$, $5k$ and $20k$ images, respectively. We trained on the training images (that contain humans) and report results on the validation set for our comparison study and on the test set for comparing to state-of-the-art models. 
Most COCO images contain only a few annotated humans. To account for this class imbalance, we down-weight the log-probability term in Equation~\ref{eq:hungarian} by a factor of 10 for all non-objects (we used the same re-weighing for Macaque Pose).

\subsection*{MacaquePose:}

The MacaquePose dataset~\cite{labuguen2020macaquepose} comprises around $13,000$ images with more than $16,000$ macaques for which the same $17$ keypoints as in COCO~\cite{lin2014microsoft} are annotated. We randomly split the dataset into $80\%$ train and $20\%$ val sets. MacaquePose does not provide any bounding box annotations, that are needed to define the scaling parameter $s$ in the OKS evaluation metric, and therefore we computed bounding boxes by fitting boxes around the provided segmentation masks.

\subsection*{Inference speed analysis:}

To analyze the inference runtime performance of the methods, we resize all images of the COCO-validation dataset to $512 \times 512$ and run inference on them. The runtime analysis is carried out on one Nvidia TITAN RTX with a batch size of 1. We calculate the average runtime for all images that contain the same amount of people and show the inference speed in terms of frames per second and dependent of the number of persons.

\subsection*{Evaluation Metrics:}

We assess the performance with the standard evaluation metric based on Object Keypoint Similarity (OKS):
\begin{equation} \label{eq:oks}
    OKS = \frac{\sum_i \exp(-d_i^2/2s^2 k_i^2) \delta(v_i>0)}{\sum_i \delta(v_i>0)}
\end{equation}
Thereby, for each keypoint $i\in \{1,2,\ldots , 17\}$, $d_i$ is the Euclidean distance between the detected keypoint and its corresponding ground truth, $v_i$ is the (boolean) visibility of the ground truth, $s$ is the object scale, $k_i$ is the labeling uncertainty (a COCO constant) and $\delta$ is $1$ for positive visibilities and zero otherwise. We calculated the average precision and recall scores: AP$_{50}$ (AP at OKS = $0.50$), AP$_{75}$, AP (the mean of AP scores at OKS = $0.50$, $0.55$, \ldots , $0.90$, $0.95$), AP$_M$ for medium objects, AP$_L$ for large objects, and AR (the mean of recalls at $OKS = 0.50, 0.55, \ldots , 0.90, 0.95$), as well as AR$_{50}$, AR$_{75}$, AR$_{M}$ and AR$_{L}$.

\subsubsection*{Inference speed analysis} To analyze the inference runtime, we resize all images of the COCO-validation dataset to 512x512 and run inference on them. The analysis was carried out on one Nvidia TITAN RTX with a batch size of 1. We calculated the runtimes averaged over all images that contain the same amount of people.

\subsection*{Implementation Details:}

We trained all (POET) models with the following coefficients in the keypoint loss: $\lambda_{abs} = 4$, $\lambda_{vis} = 0.2$, $\lambda_{ctr} = 0.5$ and $\lambda_{deltas} = 0.5$ (but see Table~\ref{table_lossablations3} and ~\ref{table_lossablations}). 

We set the transformer's initial learning rate to $0.5\cdot 10^{-4}$, the backbone's to $0.5\cdot 10^{-5}$, and weight decay to $10^{-4}$~\cite{carion2020end} and train POET with AdamW~\cite{loshchilov2017decoupled}. A dropout rate of $0.1$ is applied on the transformer's weights, which are initialized with Xavier initialization~\cite{glorot2010understanding}. For the encoder, we choose ResNet50~\cite{he2016deep} with different strides $S$. Accordingly, models are called POET-R50 as well as POET-DC5-R50, when using a dilated C5 stage (which decreases the stride from 32 to 16). The replacement of a stride by a dilation in the last stage of the backbone increases the feature resolution by a factor of two, but comes with an increase in computational cost by the same factor.

During training we augmented by applying rotation uniformly drawn from $(-25, +25)$ degrees, random cropping, horizontal flipping and coarse dropout~\cite{devries2017improved} with a probability of $0.5$ each. Additionally, we resize the images such that the shortest side falls in the range $[400,800]$ and the longest side is at most $1,333$. We set the number of prediction slots $N$ to $25$, as the maximum number of keypoint annotated humans in COCO images is $13$ (we used the same number for MacaquePose).

We carried out two different sets of experiments: (1) training POET initialized from ImageNet~\cite{russakovsky2015imagenet} weights to compare with current state-of-the-art models and (2) training multiple models as well as baseline models, whereby we started with COCO keypoint challenge pretrained weights from MMPose~\cite{mmpose2020}. 

For the comparison with state-of-the-art models, we train POET-R50 with a batch size of 6 on two NVIDIA V100 GPUs (hence a total batch size of 12) for 250 epochs, with a learning rate drop by a factor of 10 after 200 epochs. One epoch takes approximately one hour in this setting.

For the comparison to baseline models, we utilized the associative embedding~\cite{newell2017associative, cheng2020higherhrnet} implementation in MMPose~\cite{mmpose2020,mmpose2020aeresnet}. To account for the high memory footprint and the long training times, we restrict the maximum image size to 512 during training and train POET models (POET-R50, and POET-DC5-R50) with a total batch size of 64 (50 for DC5 models; we also multiply the learning rates by a factor of 2 here) for 250 epochs, for the comparison with the baseline models. The baseline models were trained for 100 epochs with the default learning schedule for AE + ResNet models in MMPose~\cite{mmpose2020aeresnet}, with similar augmentation methods (without coarse dropout, but with affine transformation augmentation). We reduced the stride by removing upsampling layers. We trained the baseline models only for 100 epochs, as we initialized the AE + ResNet models from the one trained with stride 4 on COCO and the performance saturated already. 

\input{fig_arxiv2/map_evolution}

\section*{Results}

\subsection*{Qualitative Results:}

When we trained POET-R50 on COCO with the proposed loss (Eq.~\ref{eq:hungarian}) as well as the cross-validated hyperparameters, we found that class, keypoint, visibility, center and delta loss decreased (Figure~\ref{fig:losses}). When visualizing predictions of this POET-R50 model for COCO-val, we saw that POET can successfully tackle the problem of multi-instance pose estimation (Figure~\ref{fig:example_imgs}). Next we quantified the performance. 

\subsection*{Quantitative Evaluation:}

Firstly, to quantify the performance, we computed mAP over the learning period and found that it reaches high performance (Figure~\ref{fig:mAPevolution}). Next, we compare our results to state-of-the-art methods on COCO test-dev (Table~\ref{table_results_sota}). We split the methods into top-down and bottom-up approaches and report numbers without multi-scale testing or extra training data in order to have a fair comparison. We find that POET-R50 performs competitively with other bottom-up methods for large humans ($AP_L$), but has lower performance for small/medium humans. 

\input{fig_arxiv2/poet_vs_baseline}

\begin{figure*}
  \centering
  \includegraphics[width=\linewidth]{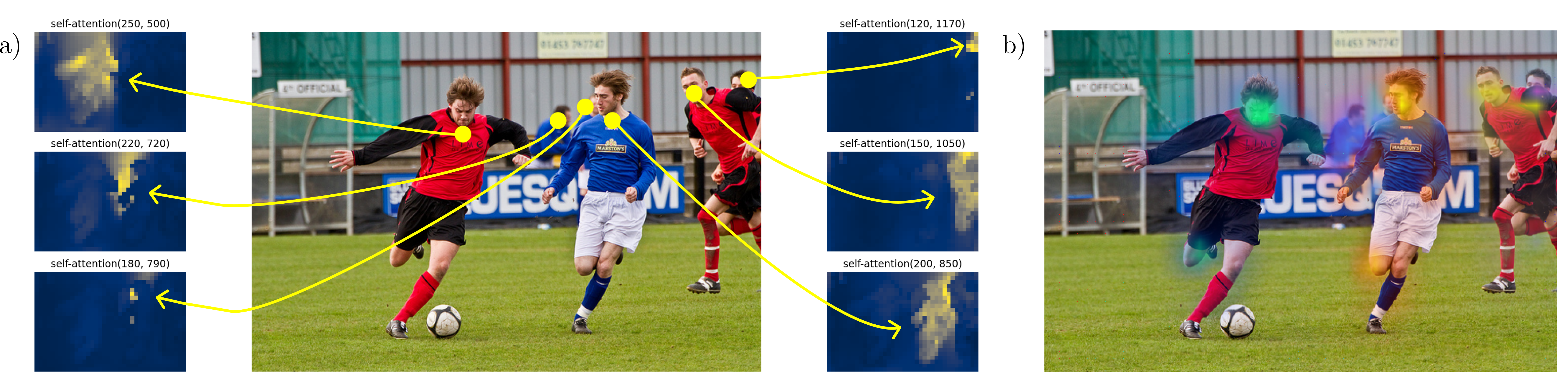}
  \caption{{\bf a)} The encoder self-attention maps for six reference points, highlighted in yellow, are shown. Pixels belonging to individuals locally attend to other pixels belonging to the same individual. Thus, the encoder learnt well to distinguish between different instances. {\bf b)} The decoder cross-attention scores for predicted individuals are highlighted for six object queries. Each focuses on a different person; for larger persons the attention is higher on extremities (such as the knee or shoulder) and especially the face while for smaller persons it is more distributed over the whole body. The image is from the COCO validation set.}
    \label{fig:attention}
\end{figure*}

We reasoned that this is due to larger stride for the encoders in all strong methods (e.g. $\geq 4$, see Table\ref{table_results_sota}), which contributes to inferior spatial resolution at the input to the transformer. Transformers scale quadratically $\mathcal{O}((H\cdot W/S^2)^2)$ in the input dimensions, and thus increasing the stride is costly. In order to demonstrate the powerful potential of our method, in comparison to previous methods, we next compare POET to baseline models with ResNet backbones (and varying strides). We chose associative embedding (AE)~\cite{newell2017associative,mmpose2020} as the model to compare to, as this method was proven to be a strong bottom-up method and currently is state-of-the-art when applied with the high resolution backbone HigherHRNet~\cite{cheng2020higherhrnet}. We created baseline models that were trained with the same pretrained ResNet backbones, input image sizes and similar data augmentation. We varied the overall stride of the ResNet backbones for AE from $4$ to up to $32$, to assure that both methods receive the same feature dimensions as input. POET outperforms the baseline methods (with the same stride) by a large margin and (for stride 16) is better than the baseline models even with stride $4$, which provides evidence that the transformer head is suitable to learn multiple poses in an image, even from low-resolution feature maps (Figure~\ref{fig:results_baseline}). Future work could focus on more efficient models to leverage smaller strides.

\subsection*{Analysis:}

\subsubsection*{Transformer Attentions}

In order to better understand the roles of the encoder and decoder in the transformer architecture for tackling multi-instance pose estimation, we visualized the attention maps. We found that the transformer's encoder attends locally to each individual while the decoder seems to attend to salient parts of each individual  (Figure~\ref{fig:attention}a,b). 

\subsubsection*{Decoder Analysis}

Next, we analyzed the role of the transformer decoder's depth. We found that the average performance stabilizes after 3-5 decoder layers (Figure~\ref{fig:mAPacrosstrafo}).

\input{fig_arxiv2/mapdecoder}

\subsubsection*{Loss Ablations}

Given that our loss (Eq.~\ref{eq:pose}) depends on four parts, we sought to identify which terms are most relevant for the performance. In the following ablation experiments we refer to $\lambda_{abs}\|V_i \circ A_i - V_i \circ \widehat{A}_{\sigma(i)}\|_1$ as \textit{$L_{abs}$} and to $\lambda_{deltas}\|V_i \circ Z_i - V_i \circ \widehat{Z}_{\sigma(i)}\|_1$ as \textit{$L_{deltas}$}. We interrogated the relationship between the loss on absolute coordinates (\textit{$L_{abs}$}) and the center (\textit{$L_{ctrs}$}) and deltas loss (\textit{$L_{deltas}$}). As their weight is already small compared to \textit{$L_{abs}$} for the best parameters we found (Figure~\ref{fig:losses}), we first set both the deltas loss part and the center loss part to $0$ and thereby giving the model freedom for predicting centers and offsets. We found that the performance stays the same. Secondly, we increased $\lambda_{ctrs}$ and $\lambda_{deltas}$ so that the loss magnitudes are comparable to \textit{$L_{abs}$}. Performance drops when center and delta loss parts are weighted higher (Table~\ref{table_lossablations3}). Next, we tested if using losses on centers and deltas alone are sufficient to reach high performance. To run more experiments, we only trained for 50 epochs. We first defined an extended version of \textit{$L_{deltas}$} in which, additionally to the distances between offsets, the L1 distance between the predicted center and the ground truth center is included (\textit{$L_{deltas\&ctr}$}).  Including this term improved the performance when the absolute error loss part was ablated (Table~\ref{table_lossablations}).
To show that the same results hold also when training for the full $250$ epochs, we took the best model trained without the absolute error loss part and compared its performance after 250 epochs to POET trained with the full loss. We conclude that the usage of the error on absolute coordinates \textit{$L_{abs}$} is crucial for the performance of POET and that using separate loss terms for centers and offsets, respectively, does not work for various parameter combinations. Thus, including $L_{abs}$ in the pose loss is crucial.

\begin{figure*}
  \centering
  \includegraphics[width=\linewidth]{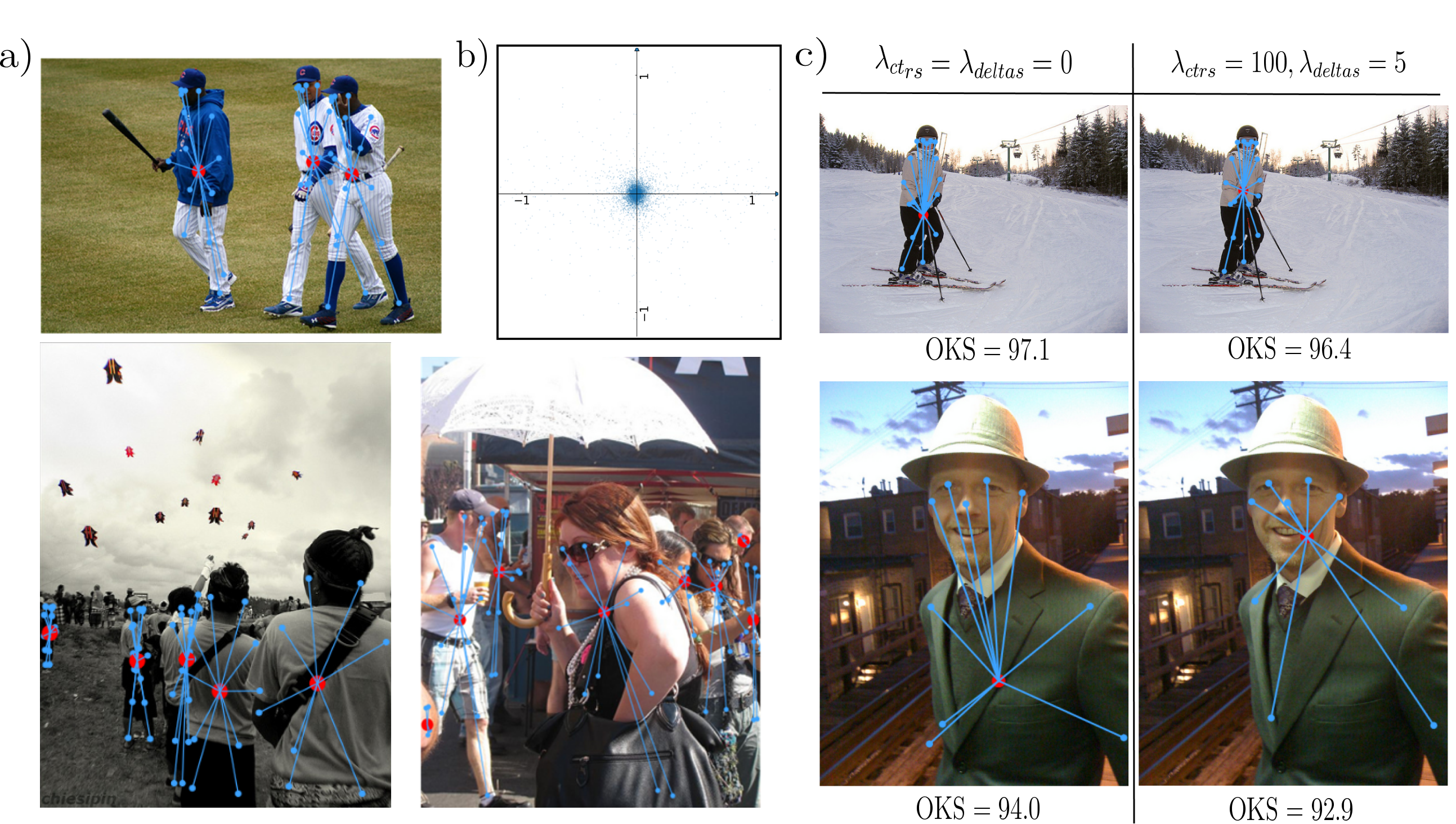}
  \caption{Analysis of learned centers. (a) Predicted centers (red) and relative displacements per bodypart as blue vectors for multiple individuals on example images from COCO val. (b) The scatter plot shows the relative offset of predicted center vs. ground truth center (normalized by bounding box width/height). (c) Predictions of two models, differently constrained on predicting the defined ground truth center (see Table~\ref{table_lossablations3}). The model without constraints ($\lambda_{ctrs}=\lambda_{deltas}=0$) learns centers that are far apart from the center of mass of visible keypoints and achieves also higher performance in terms of OKS.}
    \label{fig:center_ana}
\end{figure*}

\subsubsection*{Learned Centers}
As shown in the last section, POET models with weaker constraints on centers and deltas show the best performance. This raises the questions, where are the predicted centers? By making the body part locations dependent on a center, every joint position carries the information about the person instance it belongs to (Figure~\ref{fig:idea}b). Transformers are particularly well suited at modeling interactions and relations between data points and representations. By encoding the human pose into centers and offsets the architecture can benefit from reasoning about person instances and their poses at the same time.

However, the {\it center} is  rather complex, as it strongly depends on the visible keypoints, and the posture. Therefore, the model needs not only to understand the geometry of a human body, but also its appearance in a given image, as often only parts of the body can be seen (Figure~\ref{fig:center_ana}a). In the top image all persons can be fully visible, and POET predicts the center in the middle of the body. In the lower images, persons of different poses, scales and occlusions require a very flexible interpretation of the center. This suggests that POET predicts the centers well. Indeed, the learned centers are very close to the GT centers (Figure~\ref{fig:center_ana}b).

However, this pose representation is very flexible. One can vary the center and deltas loss weight strongly with relatively little change on quantitative (Table~\ref{table_lossablations3}) and qualitative results while strongly impacting the center location (Figure~\ref{fig:center_ana}c).

\subsubsection*{Visibility}
 
\begin{figure}
  \centering
  \includegraphics[width=\linewidth]{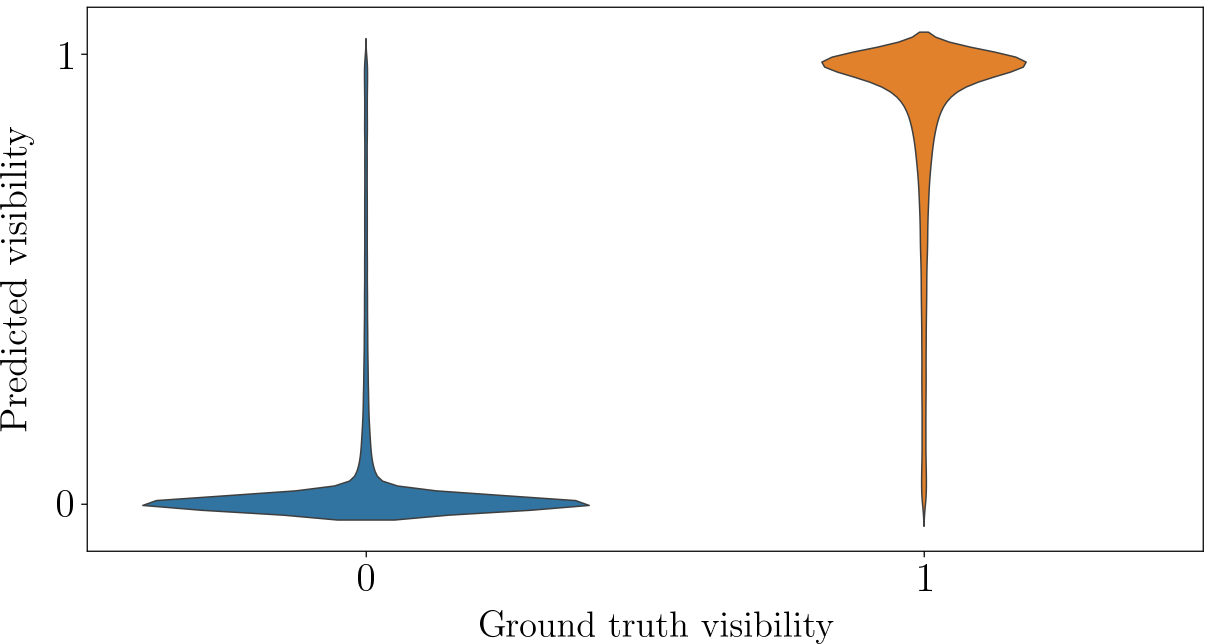}
  \caption{ Violin plot of POET-R50 predicted visibility vs. COCO ground truth visibility for all images and annotated humans in COCO-val. On COCO-val, POET-R50 predicts $94.84\%$ of all $123,359$ non-visible keypoints with a visibility score smaller than $0.5$, and $91.09\%$ of all the $59,850$ visible keypoints with a visibility confidence larger than $0.5$.}
    \label{fig:violin_ctr}
\end{figure}

In applications, it is often important not only to estimate the location of body parts, but also to ascertain if a particular bodypart is occluded~\cite{mathis2018deeplabcut}. However, the metrics used for the COCO keypoint detection challenge do not take visibility into account (i.e. false-alarm predictions are not penalized in Eq.~\ref{eq:oks}). Our loss formulation (Eq.~\ref{eq:bipartite}) allows the model to learn the visibility of each keypoint together with its location. We found that POET accurately predicts the corresponding visibility for each predicted body part (Figure~\ref{fig:violin_ctr}).

\begin{figure*}
  \centering
  \includegraphics[width=\linewidth]{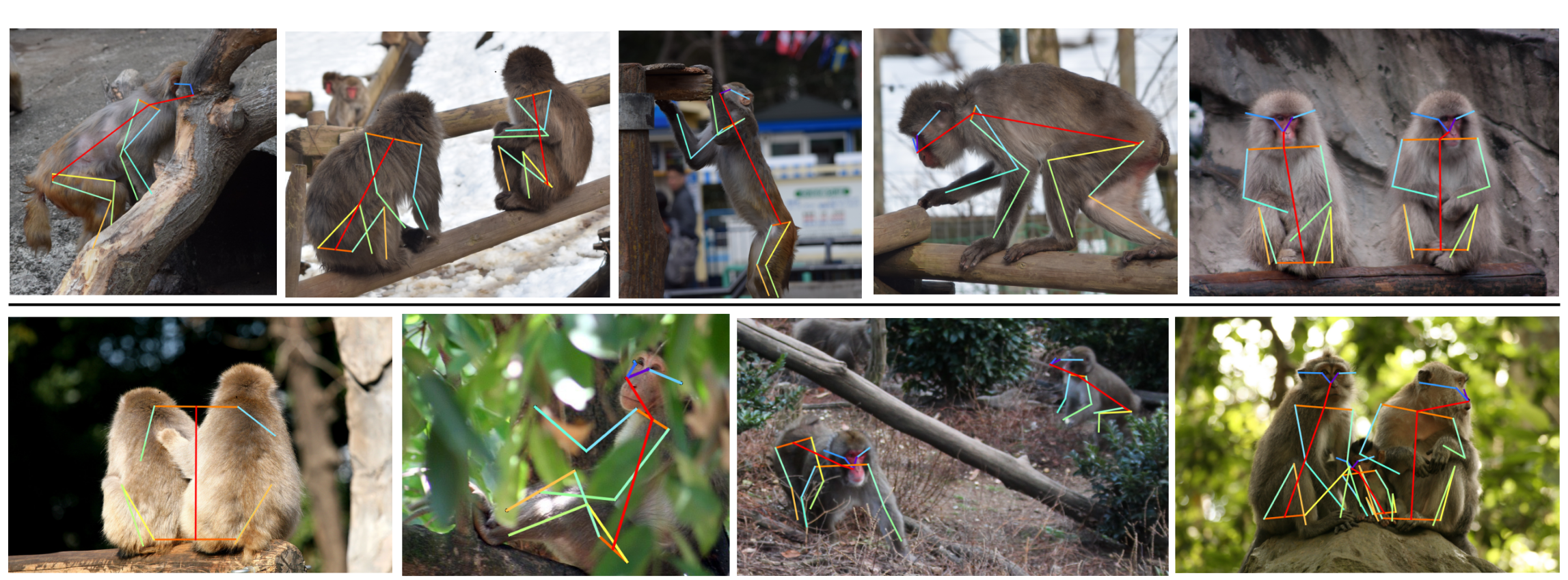}
  \caption{Example predictions on MacaquePose validation set. Top row: examples with good performance. Bottom row: examples with errors.}
    \label{fig:example_imgs_mp}
\end{figure*}

\subsection*{Inference speed} is another important characteristic of pose estimation algorithms for applications~\cite{kane2020real}. We compared the inference speed of POET with four well-known multi-person pose estimation models: the bottom-up methods OpenPose~\cite{cao2017realtime} and Associative Embedding~\cite{newell2017associative} in combination with R50~\cite{he2016deep} and HRNet backbones~\cite{sun2019deep, wang2020deep} as well as a top-down method consisting of a Mask R-CNN~\cite{he2017mask} for bounding box detection and a R50-SimpleBaseline~\cite{xiao2018simple} network for pose estimation. 

As expected for top-down approaches, the inference times of Mask R-CNN and R50-SimpleBaseline is proportional to the number of proposals that the person detector (Mask R-CNN) extracts (Figure~\ref{fig:speed_fps_wo}). A similar trend can be observed for Associative Embedding with R50/HRNet backbone. In contrast, the widely-used OpenPose method is invariant to the number of people in the image. The same holds for POET that also shows slightly higher inference speeds than OpenPose (Figure~\ref{fig:speed_fps_wo}).

\subsection*{POET for animal pose estimation:}

Datasets for animal pose estimation are typically much smaller than human benchmarks~\cite{mathis2018deeplabcut,mathis2020primer}. We thus sought to test POET on the MacaquePose dataset~\cite{labuguen2020macaquepose} to evaluate if a transformer based architecture gives competitive performance on small datasets. We found that POET qualitatively performs well on MacaquePose (Figure~\ref{fig:example_imgs_mp}). We quantified the performance of different POET instances (Table~\ref{table_results_mp}). As expected, the version with a overall stride 16 (POET-DC5-R50) outperformed the stride 32 model (POET-R50). POET outperforms associative embedding with the same backbone but a stride of four. 

As MacaquePose has the same keypoints as COCO, we experimented also with transfer learning by initializing POET with the weights of the POET-R50 model that was first pre-trained on the COCO keypoint challenge (see POET-R50 in Figure~\ref{fig:results_baseline}). We found that using this pre-trained model further boosts the performance on MacaquePose. Perhaps as expected, this model also learned much quicker, while quickly unlearning COCO (Figure~\ref{fig:mAPevolution_MP}).

\input{fig_arxiv2/map_evolution_MP}

\section*{Conclusions}

We presented POET, a pose estimation method based on a convolutional encoder, transformers and bipartite matching loss for direct set prediction. Our approach achieves strong results on COCO keypoint challenge and is the first that is end-to-end trainable. POET builds on DETR~\cite{carion2020end}, which tackled object recognition and panoptic segmentation, with transformers, and extends this framework by introducing a novel loss function that, combined with a suitable pose representation, can successfully tackle the problem of multi-instance pose estimation. At the current stage, POET does not achieve state-of-the-art performance, but we hope that it will inspire future research.

\section*{Acknowledgements}

We are grateful to Steffen Schneider, Shaokai Ye, Mu Zhou, and Mackenzie Mathis for discussions as well as Axel Bisi, Alberto Chiappa, Alessandro Marin Vargas, Nicholas Robertson, Lazar Stojkovic, Liza Kozlova, Haozhe Qi, and Aristotelis Economides for comments on this manuscript.

\begin{table*}[ht!]
\centering
\caption{Effect of center loss and deltas loss components on AP. We trained a model turning off the loss on the centers and the deltas (first row). The second row shows our baseline model with all loss terms. The maximum input image size is kept at $512$, the number of parameters is $41.3$M and class loss as well as visibility loss are used by all models.}
\label{table_lossablations3} \footnotesize
\begin{tabular}{l|ccc|cccc}
\toprule
Method & $\lambda_{abs}$ & $\lambda_{ctrs}$ & $\lambda_{deltas}$ & AP & AP$_{M}$ & AP$_{L}$ & AR \\
\toprule
\midrule
POET-R50 & 4.0 & 0.0 & 0.0 & 48.5 & 33.4 & 68.1 & 53.8 \\
POET-R50 & 4.0 & 0.5 & 0.5 & 48.5 & 33.4 & 68.3 & 53.7 \\
POET-R50 & 4.0 & 100.0 & 0.5 & 48.2 & 33.1 & 67.9 & 53.0 \\
POET-R50 & 4.0 & 0.5 & 5.0 & 47.9 & 32.7 & 67.6 & 52.9 \\
POET-R50 & 4.0 & 100.0 & 5.0 & 47.0 & 31.9 & 66.6 & 52.0 \\
\bottomrule
\end{tabular}
\end{table*}

\begin{table*}[ht!]
\centering
\vspace{15pt}
\caption{Effect of pose loss components on AP. We trained several models turning off the loss on the absolute coordinates and solely training with the center and offset loss parts. Additionally, we included the center into the L1 delta loss for some models as we saw strong performance boosts. We observe that after 50 epochs no model can reach the performance of POET trained {\bf with} the L1 loss on absolute coordinates (last row). Additionally, the best performing model after 50 epochs remains worse also after 250 epochs. The maximum input image size is kept at $512$, the number of parameters is $41.3$M and class loss as well as visibility loss are used by all models.}
\label{table_lossablations} \footnotesize
\begin{tabular}{l|ccc|c|c}
\toprule
Method & $\lambda_{abs}$ & $\lambda_{deltas}$ & $L_{deltas\&ctrs}$ & AP (after 50 epochs) & AP (after 250 epochs) \\
\toprule
\midrule
POET-R50 & - & 5.0 & - & 8.4 & - \\
POET-R50 & - & 5.0 & \checkmark & 18.0 & - \\
POET-R50 & - & 20.0 & - & 11.1 & - \\
POET-R50 & - & 20.0 & \checkmark & 20.0 & 35.5 \\
POET-R50 & - & 50.0 & - & 9.3 & - \\
POET-R50 & - & 50.0 & \checkmark & 19.0 & - \\
POET-R50 & - & 500.0 & - & 2.1 & - \\
POET-R50 & - & 500.0 & \checkmark & 12.0 & - \\
\midrule
POET-R50 & - & 0.5 & - & 2.2 & - \\
POET-R50 & - & 0.5 & \checkmark & 18.7 & - \\
POET-R50 & 4.0 & 0.5 & - & {\bf 32.2} & {\bf 48.5} \\
\bottomrule
\end{tabular}
\end{table*}

\begin{table*}[ht!]
\centering
\vspace{15pt}
\caption{Performance on MacaquePose~\cite{labuguen2020macaquepose} for various POET models. Using a CNN with a smaller stride improves performance, and initializing weights from a pre-trained model on the COCO keypoint challenge gives a strong boost. We compare POET to a ResNet-50 + associative embedding (AE)~\cite{newell2017associative, mmpose2020aeresnet} baseline trained for 300 epochs.}
\label{table_results_mp} \footnotesize
\begin{tabular}{l|c|ccc|ccc}
\toprule
Method & Input Size / Stride &  AP & AP$_{50}$ & AP$_{75}$  & AR & AR$_{50}$ & AR$_{75}$ \\
\toprule
\midrule
AE + R50 & 512 / 4 & 52.5 & 81.2 & 50.8 & 68.6 & 88.6 & 68.6 \\
AE + R50 (COCO pre-trained) & 512 / 4 & 75.4 & 95.4 & 81.4 & 83.8 & 97.2 & 88.7 \\
\midrule
POET-R50 & 512 / 32  & 65.5 & 92.8 & 75.1 & 77.9 & 97.0 & 87.0 \\
POET-DC5-R50 & 512 / 16 & 68.2 & 94.5 & 78.1 & 78.2 & 97.1 & 87.0 \\
POET-R50 (COCO pre-trained) & 512 / 32 & 75.5 & 96.0 & 85.2 & 82.0 & 97.8 & 90.1 \\
POET-DC5-R50 (COCO pre-trained) & 512 / 16 & 77.1 & 96.1 & 86.7 & 83.6 & 97.9 & 91.3 \\
\bottomrule
\end{tabular}
\end{table*}

\newpage

\section*{References}

\input{output.bbl}
\end{document}

%% file: fig_arxiv2/inference_time_fps_woDef.tex
\pgfplotsset{compat = newest, legend style={at={(0.6,0.64)},anchor=west}}

\pgfkeys{/pgfplots/tuftelike/.style={
  semithick,
  tick style={major tick length=4pt,semithick,black},
  separate axis lines,
  axis x line*=bottom,
  axis x line shift=10pt,
  xlabel shift=10pt,
  axis y line*=left,
  axis y line shift=10pt,
  ylabel shift=10pt}}

\begin{figure}[ht!]
    \centering
    \begin{tikzpicture}[
        trim left=-0.5in, 
        trim right=\columnwidth-0.2in,
        scale=0.95
    ]
        \begin{axis}[
            tuftelike,
            enlarge x limits=false,
            ylabel near ticks, 
            ylabel shift={-1pt},
            xlabel near ticks, 
            xlabel shift={-1pt},
            xmin = 1, 
            xmax = 13,
            ymin = 0, 
            ymax = 45, 
            xtick distance = 1,
            ytick distance = 5,
            grid = both,
            minor tick num = 0, 
            major grid style = {lightgray!20},
            minor grid style = {lightgray!20}, 
            point meta=y,
            xlabel = Number of people per image,
            ylabel = Frames per second (FPS),
            legend style={at={(0.57,0.45)}, font=\footnotesize}
        ]
            \addplot[red, mark=*] coordinates {
                (1, 3.561618762541135)
                (2, 2.130148293494635)
                (3, 1.5603331090078756)
                (4, 1.2195347143113653)
                (5, 0.9965550819620681)
                (6, 0.8613629219204025)
                (7, 0.7304498281413723)
                (8, 0.6252930631381849)
                (9, 0.571476105829135)
                (10, 0.5155409543012394)
                (11, 0.47948916270415143)
                (12, 0.44059178553991807)
                (13, 0.4207557965374353)
            };
            \addplot[orange, mark=*] coordinates {
                (1, 2.4830317643520403)
                (2, 1.504985944620749)
                (3, 1.2116728838275648)
                (4, 0.8663104959125782)
                (5, 0.6505787662915865)
                (6, 0.7042021400552173)
                (7, 0.6108158016720991)
                (8, 0.47860610062843706)
                (9, 0.39791818677821483)
                (10, 0.4306742716135787)
                (11, 0.36183014268566516)
                (12, 0.3830569059017311)
                (13, 0.3339146473375713)
            };
            \addplot[curveblue, mark=*] coordinates {
                (1, 34.57267830605305)
                (2, 33.911792328667204)
                (3, 32.72986145105496)
                (4, 33.43400495803411)
                (5, 33.175460755681875)
                (6, 32.6024352611762)
                (7, 33.533282763866644)
                (8, 32.15229945033407)
                (9, 33.06696928193733)
                (10, 33.73590085312784)
                (11, 32.31003238021444)
                (12, 34.07024644579021)
                (13, 32.39021537340004)
            };
            \addplot[black, mark=*] coordinates {
                (1, 11.7614456084405)
                (2, 9.987691561651234)
                (3, 8.595600314846394)
                (4, 7.662885414955848)
                (5, 6.885541838049584)
                (6, 6.268496199868848)
                (7, 5.669892879370095)
                (8, 5.180215879560219)
                (9, 4.9522922822965665)
                (10, 4.571543613828987)
                (11, 4.381253807930808)
                (12, 4.033573867130455)
                (13, 3.8708920732491934)
            };
            \addplot[black, dashed, mark=*] coordinates {
                (1, 54.23058256325218)
                (2, 30.966529052816835)
                (3, 20.963116762560443)
                (4, 16.37155885972632)
                (5, 13.194047182311849)
                (6, 11.061002519196045)
                (7, 9.402310060155461)
                (8, 8.217889127064145)
                (9, 7.4367665778982515)
                (10, 6.906813178115921)
                (11, 6.186376804685854)
                (12, 5.787355552632934)
                (13, 5.366552195341648)
            };
            \addplot[curvegreen, mark=*] coordinates {
                (1, 40.08639727309373)
                (2, 40.16501052380964)
                (3, 39.98034518281845)
                (4, 40.44937310531113)
                (5, 40.26276648495852)
                (6, 40.17337367271775)
                (7, 39.45683535865183)
                (8, 40.02896316770718)
                (9, 40.21653092200789)
                (10, 40.32729540394155)
                (11, 40.61750919255892)
                (12, 40.333004095100854)
                (13, 39.98716256305717)
            };
            \legend{AE+R50, AE+HRNet, OpenPose, Top-Down, Top-Down w/o bb det, \textbf{POET}}
      \end{axis}
    \end{tikzpicture}
    \caption{Inference speed comparison between AE+R50 (Bottom-Up), AE+HRNet (strong Bottom-Up), OpenPose, Mask R-CNN + R50 Simple Baseline (Top-Down), the same Top-Down baseline with provided bounding boxes, and POET. 
    POET's inference speed is invariant to the number of people in an image and outperforms previous multi-instance pose estimation methods. Tests were performed on the COCO-val set with a batch size of one on a single Nvidia TITAN RTX.}
    \label{fig:speed_fps_wo}
\end{figure}

%% file: fig_arxiv2/map_evolution.tex
\pgfplotsset{compat = newest, legend style={at={(0.65,0.8)},anchor=west}}

\pgfkeys{/pgfplots/tuftelike/.style={
  thick,
  tick style={major tick length=4pt,semithick,black},
  separate axis lines,
  axis x line*=bottom,
  axis x line shift=10pt,
  xlabel shift=10pt,
  axis y line*=left,
  axis y line shift=10pt,
  ylabel shift=10pt}}

\begin{figure}[ht]
    \centering
    \begin{tikzpicture}[
        trim left=-0.7in, 
        trim right=\columnwidth,
        scale=0.8
    ]
        \begin{axis}[
            tuftelike,
            enlarge x limits=false,
            ylabel near ticks, 
            ylabel shift={-10pt},
            xlabel near ticks, 
            xlabel shift={-1pt},
            xmin = 0, 
            xmax = 250,
            ymin = 0, 
            ymax = 60, 
            xtick distance = 50,
            ytick distance = 10,
            grid = both,
            minor tick num = 0, 
            major grid style = {lightgray!10},
            minor grid style = {lightgray!10}, 
            point meta=y,
            xlabel=Training epochs,
            ylabel=COCO mAP performance ($\%$),
        ]
            \addplot[curvemap] coordinates {
                (0, 0.0)
                (1, 0.0)
                (2, 0.2)
                (3, 0.5)
                (4, 1.0)
                (5, 1.7999999999999998)
                (6, 1.5)
                (7, 4.0)
                (8, 4.8)
                (9, 6.7)
                (10, 7.7)
                (11, 7.7)
                (12, 8.4)
                (13, 11.1)
                (14, 12.9)
                (15, 11.1)
                (16, 15.1)
                (17, 12.1)
                (18, 17.1)
                (19, 14.499999999999998)
                (20, 20.7)
                (21, 18.7)
                (22, 19.6)
                (23, 22.2)
                (24, 22.3)
                (25, 21.3)
                (26, 22.2)
                (27, 22.1)
                (28, 23.599999999999998)
                (29, 18.9)
                (30, 26.700000000000003)
                (31, 26.8)
                (32, 25.4)
                (33, 27.200000000000003)
                (34, 22.0)
                (35, 28.4)
                (36, 27.800000000000004)
                (37, 28.000000000000004)
                (38, 30.7)
                (39, 29.599999999999998)
                (40, 31.4)
                (41, 30.599999999999998)
                (42, 30.0)
                (43, 30.7)
                (44, 31.8)
                (45, 30.099999999999998)
                (46, 30.599999999999998)
                (47, 29.5)
                (48, 30.3)
                (49, 31.6)
                (50, 32.1)
                (51, 33.7)
                (52, 31.5)
                (53, 34.699999999999996)
                (54, 34.2)
                (55, 34.2)
                (56, 33.300000000000004)
                (57, 32.4)
                (58, 33.4)
                (59, 32.7)
                (60, 35.099999999999994)
                (61, 36.9)
                (62, 37.8)
                (63, 35.699999999999996)
                (64, 35.699999999999996)
                (65, 37.4)
                (66, 35.8)
                (67, 35.6)
                (68, 32.9)
                (69, 37.5)
                (70, 36.199999999999996)
                (71, 33.1)
                (72, 38.2)
                (73, 38.7)
                (74, 35.3)
                (75, 37.6)
                (76, 36.3)
                (77, 37.3)
                (78, 38.0)
                (79, 39.0)
                (80, 36.9)
                (81, 40.2)
                (82, 41.0)
                (83, 40.1)
                (84, 38.3)
                (85, 39.2)
                (86, 40.5)
                (87, 39.4)
                (88, 39.800000000000004)
                (89, 40.0)
                (90, 36.5)
                (91, 40.699999999999996)
                (92, 39.0)
                (93, 40.400000000000006)
                (94, 39.7)
                (95, 41.099999999999994)
                (96, 42.1)
                (97, 42.199999999999996)
                (98, 41.0)
                (99, 41.4)
                (100, 41.5)
                (101, 36.8)
                (102, 42.5)
                (103, 41.9)
                (104, 40.699999999999996)
                (105, 41.9)
                (106, 41.199999999999996)
                (107, 41.099999999999994)
                (108, 41.199999999999996)
                (109, 41.9)
                (110, 39.800000000000004)
                (111, 41.3)
                (112, 42.3)
                (113, 42.6)
                (114, 41.8)
                (115, 42.699999999999996)
                (116, 44.9)
                (117, 42.9)
                (118, 43.7)
                (119, 42.1)
                (120, 41.9)
                (121, 42.199999999999996)
                (122, 43.5)
                (123, 41.199999999999996)
                (124, 43.0)
                (125, 43.9)
                (126, 44.2)
                (127, 42.1)
                (128, 43.4)
                (129, 44.2)
                (130, 43.5)
                (131, 44.5)
                (132, 44.0)
                (133, 44.1)
                (134, 44.5)
                (135, 44.7)
                (136, 44.5)
                (137, 45.2)
                (138, 42.4)
                (139, 45.0)
                (140, 43.9)
                (141, 43.9)
                (142, 44.0)
                (143, 42.9)
                (144, 44.5)
                (145, 45.7)
                (146, 44.4)
                (147, 44.7)
                (148, 44.7)
                (149, 45.300000000000004)
                (150, 45.4)
                (151, 44.6)
                (152, 44.800000000000004)
                (153, 45.1)
                (154, 45.1)
                (155, 45.6)
                (156, 45.300000000000004)
                (157, 44.2)
                (158, 46.800000000000004)
                (159, 46.9)
                (160, 46.0)
                (161, 46.6)
                (162, 46.1)
                (163, 41.3)
                (164, 45.7)
                (165, 41.4)
                (166, 46.0)
                (167, 46.400000000000006)
                (168, 42.5)
                (169, 46.1)
                (170, 47.0)
                (171, 46.7)
                (172, 45.300000000000004)
                (173, 44.9)
                (174, 47.0)
                (175, 44.800000000000004)
                (176, 46.0)
                (177, 46.400000000000006)
                (178, 48.5)
                (179, 47.199999999999996)
                (180, 47.0)
                (181, 47.0)
                (182, 46.9)
                (183, 46.7)
                (184, 47.599999999999994)
                (185, 48.1)
                (186, 46.0)
                (187, 47.5)
                (188, 47.099999999999994)
                (189, 46.5)
                (190, 47.599999999999994)
                (191, 45.4)
                (192, 45.7)
                (193, 46.5)
                (194, 46.300000000000004)
                (195, 47.699999999999996)
                (196, 45.4)
                (197, 48.1)
                (198, 45.9)
                (199, 47.4)
                (200, 53.2)
                (201, 53.1)
                (202, 53.400000000000006)
                (203, 53.1)
                (204, 53.5)
                (205, 53.300000000000004)
                (206, 53.300000000000004)
                (207, 53.400000000000006)
                (208, 53.1)
                (209, 53.900000000000006)
                (210, 54.1)
                (211, 54.2)
                (212, 54.0)
                (213, 54.0)
                (214, 53.6)
                (215, 53.800000000000004)
                (216, 53.900000000000006)
                (217, 54.0)
                (218, 54.2)
                (219, 54.2)
                (220, 54.1)
                (221, 54.1)
                (222, 53.800000000000004)
                (223, 54.2)
                (224, 54.0)
                (225, 53.7)
                (226, 54.50000000000001)
                (227, 53.800000000000004)
                (228, 54.0)
                (229, 54.2)
                (230, 54.400000000000006)
                (231, 53.5)
                (232, 54.2)
                (233, 54.0)
                (234, 54.2)
                (235, 54.400000000000006)
                (236, 53.400000000000006)
                (237, 53.2)
                (238, 53.800000000000004)
                (239, 53.900000000000006)
                (240, 53.800000000000004)
                (241, 53.800000000000004)
                (242, 54.2)
                (243, 53.900000000000006)
                (244, 54.2)
                (245, 54.1)
                (246, 54.1)
                (247, 54.300000000000004)
                (248, 54.6)
                (249, 54.50000000000001)
            };
      \end{axis}
    \end{tikzpicture}
    \caption{Evolution of mAP on COCO validation set for POET-R50 trained for 250 epochs. Learning rate was dropped after 200. Loss parameters: $\lambda_{abs} = 4$, $\lambda_{vis} = 0.2$, $\lambda_{ctr} = 0.5$ and $\lambda_{deltas} = 0.5$.}
    \label{fig:mAPevolution}
\end{figure}

%% file: fig_arxiv2/poet_vs_baseline.tex
\pgfplotsset{compat = newest, legend style={at={(0.65,0.8)},anchor=west}}
\pgfkeys{/pgfplots/tuftelike/.style={
  semithick,
  tick style={major tick length=4pt,semithick,black},
  separate axis lines,
  axis x line*=bottom,
  axis x line shift=10pt,
  xlabel shift=10pt,
  axis y line*=left,
  axis y line shift=10pt,
  ylabel shift=10pt}}

\begin{figure}[t]
    \centering
    \begin{tikzpicture}[
        trim left=-0.6in, 
        trim right=\columnwidth-0.1in,
        scale=0.9
    ]
        \begin{axis}[
            tuftelike,
            ylabel near ticks, 
            ylabel shift={-10pt},
            xlabel near ticks, 
            xlabel shift={-1pt},
            enlarge x limits=false,
            xmin = 4, 
            xmax = 32,
            ymin = 0, 
            ymax = 100, 
            xtick distance = 4,
            ytick distance = 10,
            grid = both,
            minor tick num = 0, 
            major grid style = {lightgray!10},
            minor grid style = {lightgray!10}, 
            point meta=y,
            xlabel=CNN stride,
            ylabel=COCO mAP performance ($\%$),
        ]
            \addplot[curveblue, mark=*] coordinates {
              (16, 52.8)
              (32, 48.5)
            };
            \addplot[mark=*] coordinates {
              (4, 46.6)
              (8, 24.9)
              (16, 10.5)
              (32, 0.9)
            };
            \legend{Ours, Baseline}
      \end{axis}
    \end{tikzpicture}
    \caption{POET vs. the fully convolutional associative embedding method performance as a function of the stride of the encoder on the COCO keypoint estimation task. Our method achieves strong performance despite lower resolution features.}
    \label{fig:results_baseline}
\end{figure}

%% file: fig_arxiv2/mapdecoder.tex
\pgfplotsset{compat = newest, legend style={at={(0.65,0.17)},anchor=west}}

\pgfkeys{/pgfplots/tuftelike/.style={
  semithick,
  tick style={major tick length=4pt,semithick,black},
  separate axis lines,
  axis x line*=bottom,
  axis x line shift=10pt,
  xlabel shift=10pt,
  axis y line*=left,
  axis y line shift=10pt,
  ylabel shift=10pt}}

\begin{figure}[ht!]
    \centering
    \begin{tikzpicture}[
        trim left=-0.7in, 
        trim right=\columnwidth-0.2in,
        scale=0.8
    ]
        \begin{axis}[
            tuftelike,
            enlarge x limits=false,
            ylabel near ticks, 
            ylabel shift={-10pt},
            xlabel near ticks, 
            xlabel shift={-1pt},
            xmin = 1, 
            xmax = 6,
            ymin = 30, 
            ymax = 72, 
            xtick distance = 1,
            ytick distance = 10,
            grid = both,
            minor tick num = 0, 
            major grid style = {lightgray!20},
            minor grid style = {lightgray!20}, 
            point meta=y,
            xlabel=Transformer decoder layer,
            ylabel=COCO mAP performance ($\%$),
        ]
            \addplot[curvered, mark=*] coordinates {
              (1, 43.8)
              (2, 52.3)
              (3, 54.3)
              (4, 54.9)
              (5, 55.1)
              (6, 55.2)
            };
            \addplot[curveblue, mark=*] coordinates {
              (1, 32.3)
              (2, 40.9)
              (3, 43.1)
              (4, 43.9)
              (5, 44.4)
              (6, 44.4)
            };
            \addplot[curvegreen, mark=*] coordinates {
              (1, 59.7)
              (2, 67.6)
              (3, 69.5)
              (4, 69.9)
              (5, 69.9)
              (6, 70.0)
            };
            \legend{all, medium, large}
      \end{axis}
    \end{tikzpicture}
    \caption{Evolution of the mAP accuracy on COCO-val through decoder layers of POET-R50 illustrating that the performance saturates after four decoding layers.}
    \label{fig:mAPacrosstrafo}
\end{figure}

%% file: fig_arxiv2/map_evolution_MP.tex
\pgfplotsset{compat = newest, legend style={at={(0.65,0.8)},anchor=west}}

\pgfkeys{/pgfplots/tuftelike/.style={
  thick,
  tick style={major tick length=4pt,semithick,black},
  separate axis lines,
  axis x line*=bottom,
  axis x line shift=10pt,
  xlabel shift=10pt,
  axis y line*=left,
  axis y line shift=10pt,
  ylabel shift=10pt}}

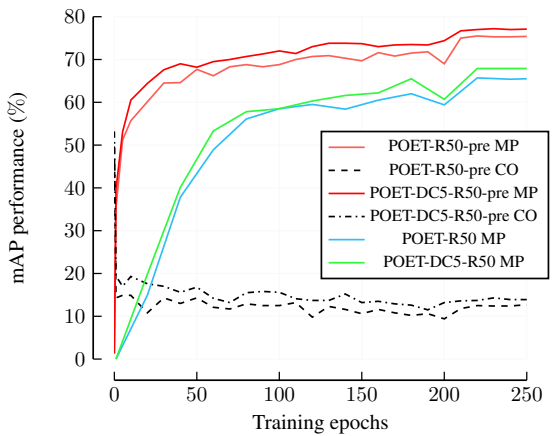
\begin{figure}[ht]
    \centering
    \begin{tikzpicture}[
        trim left=-0.7in, 
        trim right=\columnwidth,
        scale=0.8
    ]
        \begin{axis}[
            tuftelike,
            enlarge x limits=false,
            ylabel near ticks, 
            ylabel shift={-10pt},
            xlabel near ticks, 
            xlabel shift={-1pt},
            xmin = 0, 
            xmax = 250,
            ymin = 0, 
            ymax = 80,
            xtick distance = 50,
            ytick distance = 10,
            grid = both,
            minor tick num = 0, 
            major grid style = {lightgray!10},
            minor grid style = {lightgray!10}, 
            point meta=y,
            xlabel=Training epochs,
            ylabel=mAP performance ($\%$),
            legend style={at={(0.5,0.45)}, font=\footnotesize}
        ]
            \addplot[curvemap] coordinates {
                (0, 1.3)
                (1, 35.4)
                (5, 51.2)
                (10, 55.7)
                (20, 60.1)
                (30, 64.5)
                (40, 64.6)
                (50, 67.7)
                (60, 66.2)
                (70, 68.3)
                (80, 68.8)
                (90, 68.3)
                (100, 68.8)
                (110, 70.0)
                (120, 70.7)
                (130, 70.9)
                (140, 70.3)
                (150, 69.7)
                (160, 71.6)
                (170, 70.8)
                (180, 71.5)
                (190, 71.8)
                (200, 69.0)
                (210, 75.0)
                (220, 75.5)
                (230, 75.3)
                (240, 75.3)
                (250, 75.4)
            };
            
            \addplot[black, dashed] coordinates {
                (0, 49.7)
                (1, 14.3)
                (5, 14.9)
                (10, 14.9)
                (20, 10.8)
                (30, 14.3)
                (40, 13.0)
                (50, 14.3)
                (60, 12.1)
                (70, 11.7)
                (80, 12.9)
                (90, 12.5)
                (100, 12.5)
                (110, 13.2)
                (120, 9.8)
                (130, 12.3)
                (140, 11.6)
                (150, 10.6)
                (160, 11.6)
                (170, 10.8)
                (180, 10.2)
                (190, 10.6)
                (200, 9.4)
                (210, 11.8)
                (220, 12.5)
                (230, 12.4)
                (240, 12.4)
                (250, 12.7)
            };
            \addplot[red] coordinates {
                (0, 1.3)
                (1, 39.7)
                (5, 53.1)
                (10, 60.5)
                (20, 64.4)
                (30, 67.6)
                (40, 69.0)
                (50, 68.2)
                (60, 69.5)
                (70, 70.0)
                (80, 70.7)
                (90, 71.3)
                (100, 72.0)
                (110, 71.4)
                (120, 73.0)
                (130, 73.8)
                (140, 73.8)
                (150, 73.7)
                (160, 73.0)
                (170, 73.4)
                (180, 73.5)
                (190, 73.4)
                (200, 74.4)
                (210, 76.7)
                (220, 77.0)
                (230, 77.2)
                (240, 77.0)
                (250, 77.1)
            };
            
            \addplot[black, dashdotted] coordinates {
                (0, 53.2)
                (1, 19.4)
                (5, 17.0)
                (10, 19.3)
                (20, 17.6)
                (30, 17.0)
                (40, 15.6)
                (50, 16.8)
                (60, 14.2)
                (70, 13.2)
                (80, 15.5)
                (90, 15.8)
                (100, 15.6)
                (110, 14.1)
                (120, 13.7)
                (130, 13.7)
                (140, 15.2)
                (150, 13.2)
                (160, 13.5)
                (170, 12.9)
                (180, 12.6)
                (190, 11.5)
                (200, 13.2)
                (210, 13.6)
                (220, 13.7)
                (230, 14.3)
                (240, 13.9)
                (250, 13.9)
            };
                \addplot[curveblue] coordinates {
                (1, 0)
                (20, 14.9)
                (40, 37.8)
                (60, 48.9)
                (80, 56.1)
                (100, 58.5)
                (120, 59.5)
                (140, 58.4)
                (160, 60.5)
                (180, 62.0)
                (200, 59.4)
                (220, 65.7)
                (240, 65.4)
                (250, 65.5)
            };
                \addplot[curvegreen] coordinates {
                (1, 0)
                (20, 19.8)
                (40, 40.1)
                (60, 53.3)
                (80, 57.8)
                (100, 58.5)
                (120, 60.3)
                (140, 61.6)
                (160, 62.2)
                (180, 65.5)
                (200, 60.7)
                (220, 67.9)
                (240, 67.9)
                (250, 67.9)
            };
            \legend{POET-R50-pre MP, POET-R50-pre CO, POET-DC5-R50-pre MP, POET-DC5-R50-pre CO, POET-R50 MP, POET-DC5-R50 MP}
      \end{axis}
    \end{tikzpicture}
    \caption{Evolution of mAP for different POET models. Performance on the MacaquePose (MP) validation set and on COCO (CO) val are shown for the model initialized with the weights of a POET model trained on COCO (POET-pre). It learns faster and achieves higher performance than the two models trained from ImageNet initialization (POET-R50, POET-DC5-R50), while the performance on COCO decays within one epoch and plateaus (dashed lines).}
    \label{fig:mAPevolution_MP}
\end{figure}